\documentclass{ecai}
\usepackage{graphicx}
\usepackage{latexsym}
\usepackage{caption}
\usepackage{booktabs}
\usepackage{algorithm}
\usepackage{algorithmic}
\usepackage{amsmath}
\usepackage{color}

\begin{document}

\begin{frontmatter}

\title{Evaluating Explanation Methods for Vision-and-Language Navigation}

\author[A]{\fnms{Guanqi}~\snm{Chen}}
\author[B]{\fnms{Lei}~\snm{Yang}}
\author[C]{\fnms{Guanhua}~\snm{Chen}}
\author[A,B]{\fnms{Jia}~\snm{Pan}
\thanks{
% This research is partly supported by the Innovation and Technology Commission of the HKSAR Government through the InnoHK initiative.
Corresponding author is Jia Pan. Correspondence to: Guanqi Chen <chguanqi@connect.hku.hk>, Jia Pan <jpan@cs.hku.hk>.}}

\address[A]{Department of Computer Science, The University of Hong Kong, Hong Kong SAR, China}
\address[B]{Centre for Transformative Garment Production (TransGP), Hong Kong SAR, China}
\address[C]{Department of Statistics and Data Science, Southern University of Science and Technology, China}

\begin{abstract}
The ability to navigate robots with natural language instructions in an unknown environment is a crucial step for achieving embodied artificial intelligence (AI). With the improving performance of deep neural models proposed in the field of vision-and-language navigation (VLN), it is equally interesting to know what information the models utilize for their decision-making in the navigation tasks. To understand the inner workings of deep neural models, various explanation methods have been developed for promoting explainable AI (XAI). But they are mostly applied to deep neural models for image or text classification tasks and little work has been done in explaining deep neural models for VLN tasks. In this paper, we address these problems by building quantitative benchmarks to evaluate explanation methods for VLN models in terms of faithfulness. We propose a new erasure-based evaluation pipeline to measure the step-wise textual explanation in the sequential decision-making setting. We evaluate several explanation methods for two representative VLN models on two popular VLN datasets and reveal valuable findings through our experiments.
% The abstract should contain no more than 200 words.
\end{abstract}

\end{frontmatter}

\section{Introduction}

In the Vision-and-Language Navigation (VLN) task~\cite{Anderson_2018_CVPR}, the agent is instructed via natural language utterances to navigate to a specified location in a photo-realistic environment. To accomplish this task, the agent needs to understand the language instruction, perceive the environment, and align the instruction with the multi-modal information in the environment. In recent years, VLN has received increasing attention from computer vision, natural language processing, and robotics. A variety of studies have been conducted, covering aspects such as data augmentation \cite{fried2018speaker, liu2021vision}, representation learning \cite{hong2021vln, huang2019transferable}, multi-modal grounding \cite{hu2019you, qi2020object}, and action strategy learning \cite{wang2020active, wang2019reinforced}. These efforts have led to significant improvements in the navigation performance. 
With these advancements made in VLN, it is equally interesting to pursue a better understanding of how these seemingly reasonable actions are made by the state-of-the-art VLN models.  

\begin{figure}[tbp]
\centering
\includegraphics[width=0.43\textwidth]{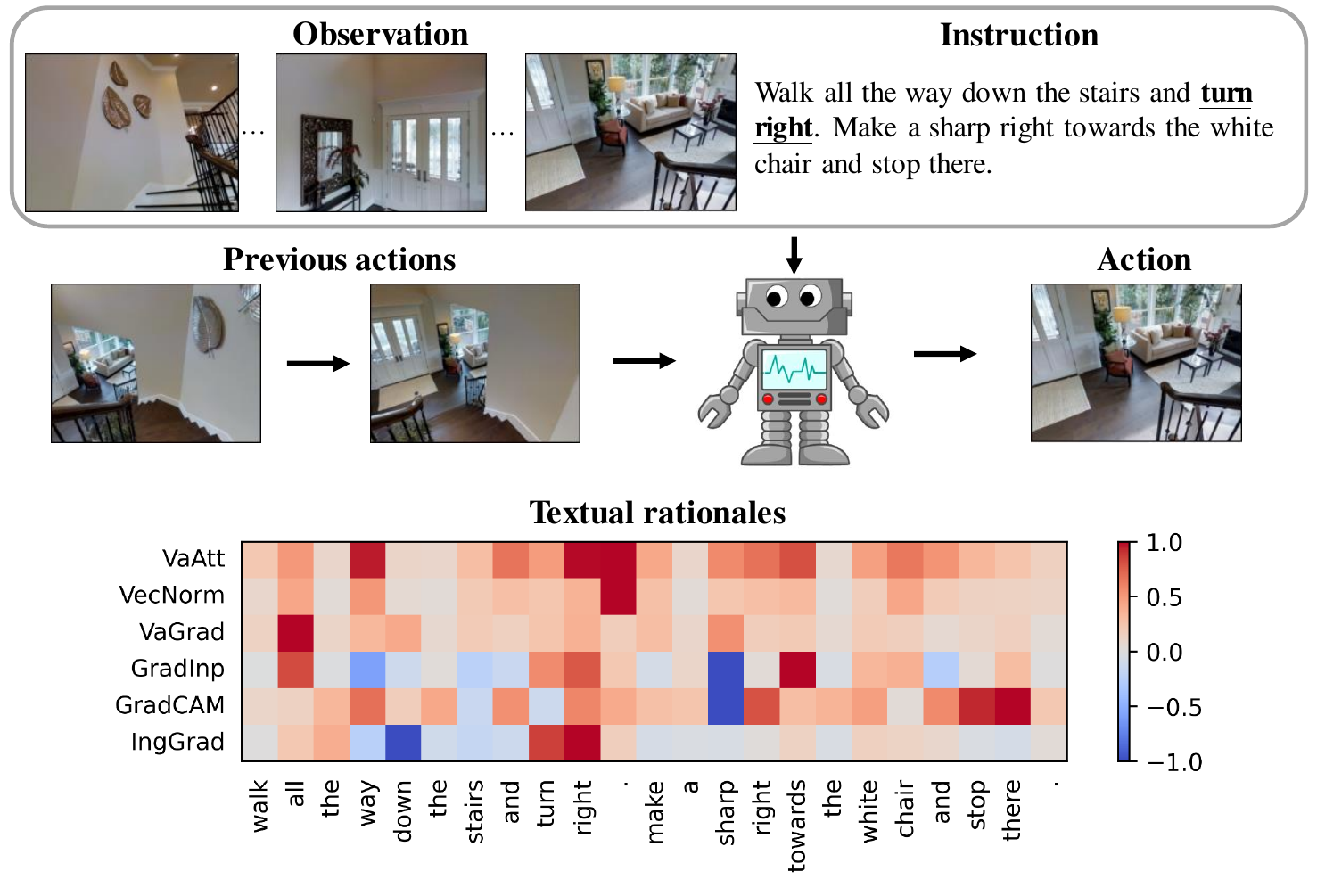}
\caption{An example from our evaluation where different explanation methods predict different textual rationales for a VLN model which makes navigation decisions based on input instruction, panoramic observation, and historical information.
The textual rationales refer to the importance scores for individual words. The patches of red and blue represent positive and negative importance scores, respectively. The bold underlined words in the gray box signify the words that human annotators rely on when making the decision. VaAtt, VecNorm, VaGrad, GradInp, GradCAM, and IngGrad are different explanation methods (see Section 2).} \label{fig1}
\end{figure}

Discovering the rationale behind decision-making processes is the primary target of explainable Artificial Intelligence (XAI) \cite{lyu2022towards, heuillet2021explainability}, an emerging field for revealing the inner workings of AI systems. In XAI, a number of works have been proposed to explain the output of deep neural networks. Since most VLN models are built on deep neural networks, these general explanation methods \cite{kobayashi2020attention, clark2019does, simonyan2013deep, denil2014extraction, selvaraju2017grad, sundararajan2017axiomatic} would be applicable to explain VLN models. However, we made an observation as shown in Fig.~\ref{fig1} that different explanation methods often yield different rationales. Therefore, it is crucial to evaluate their performance on state-of-the-art VLN models and understand the applicability of existing explanation methods.

In this paper, we establish a quantitative benchmark to evaluate explanation methods for VLN. Our benchmark specifically focuses on analyzing the textual rationales which are represented by the importance scores for individual words obtained through explanation methods. Building upon the guidelines proposed by \cite{jacovi2020towards}, we evaluate explanation methods from the perspective of faithfulness. \textit{Faithfulness} refers to how accurately the predicted explanation reflects the decision-making process of the model. Recent works~\cite{ding2021evaluating,li2020evaluating} approached this problem by comparing the output consistency between the original deep model and a proxy model. We alternatively propose an erasure-based evaluation protocol, inspired by \cite{deyoung2019eraser}, to measure the faithfulness of different methods.

Given a language instruction, a naive approach is to erase, at a certain step, the words receiving higher importance scores from the VLN models. However, this simple practice may not work for VLN models due to the sequential nature of the task. That is, VLN models make their decisions based not only on the current observations but also on historical information which includes both textual and visual clues the agent collects in the previous steps. 

For example, if the instruction for the VLN model is "Go to the living room, and then turn left, stop near the stove inside the kitchen", erasing the word "living" in the second step may not necessarily affect the decision-making process, as the information "living room" has been exposed to the model in the first step and encoded in the historical information. 
This possible flaw of naive erasure may influence the measured results, rendering it not possible to correctly reflect the true importance of the erased words for model decision-making in a specific step.

To avoid deceptive results due to naive erasure, we present a novel pipeline for assessing faithfulness through textual erasure in the VLN task. To measure the faithfulness of the explanation of step $t$ at location $loc_t$, we erase some important word tokens indicated by the explanation method from the input instruction at the starting point. Then we enforce the agent to take the same actions predicted by the original agent whose textual input is the full instruction. When the agent achieves the location $loc_t$,  we will calculate the rate of decision changes and the difference in decision probabilities at time $t$ after erasing important words as part of faithfulness metrics, which are used to evaluate whether the tokens assigned lower weights are unnecessary for the predictions. Besides, we also use a similar process to measure whether the tokens assigned with higher weights are sufficient for a model to make the original decision by preserving the important tokens rather than erasing them.

This study investigated two types of VLN models, namely the transformer-based and RNN-based models. Additionally, we explore two categories of explanation methods, i.e., the attention-based and gradient-based methods. To evaluate these explanation methods, we conducted comprehensive experiments on R2R \cite{Anderson_2018_CVPR} and RxR \cite{rxr} datasets. Through these experiments, we aim to provide a thorough, quantitative analysis of explanation methods for VLN models and explore the inner reasoning process of these VLN models.

Technical contributions of this work are summarized as follows:
\begin{enumerate}

\item We propose a new pipeline that evaluates the faithfulness of explanation methods for understanding the VLN models which make sequential actions. To the best of our knowledge, it is the first attempt at evaluating explanation methods for vision-and-language navigation. 
\item We extensively analyze the effectiveness of six commonly used explanation methods applied to two representative VLN models on two widely used VLN datasets. Our findings include: (1) Attention maps are not faithful in explaining the decision-making of a VLN model. (2) The internal reasoning processes of two representative VLN models exhibit a degree of difference from human explanations. 
\end {enumerate}

\section{Related Work}

\subsection{Model Explainability}

In recent years, the pursuit of understanding the internal mechanics of deep neural networks has garnered an increasingly substantial amount of attention and interest within the deep learning community. The complexity and black-box nature of these networks have propelled the need for elucidating the underlying decision-making processes and factors driving their predictions. To address this pressing issue, a lot of explanation methods have been introduced, each offering unique insights into the inner workings of these powerful models.

Among the diverse array of explanation methods proposed, attention-based methods and gradient-based methods have emerged as particularly noteworthy approaches. Attention-based methods \cite{clark2019does,vaswani2017attention} extract attention weights derived during the forward propagation of the model, unveiling the degree to which the model "attends to" each input vector. In accordance with this, \cite{kobayashi2020attention} develop an explanation method by integrating attention weights with the norm of input vectors. On the other hand, gradient-based explanation methods \cite{simonyan2013deep, denil2014extraction, selvaraju2017grad, sundararajan2017axiomatic} offer an alternative avenue to comprehend the model's decision-making rationale. These methods consider the gradient (or its variant) of the model output with respect to each input feature as its relevance, as gradients reflect how much a small change in the input affects the output.

In this paper, our primary focus is to undertake a comprehensive exploration of several existing explanation methods, specifically tailoring our investigation towards interpreting VLN models. Through the designed evaluation pipeline, we identify the most effective method for interpreting these models. By doing so, we hope to provide insights that can help researchers diagnose and debug their VLN models.

\subsection{Explanation Evaluation}
To evaluate explanation methods, researchers have introduced various properties that a good explanation should satisfy. The two most widely accepted properties are faithfulness and plausibility. Faithfulness refers to how accurately the explanation reflects the reasoning process behind the model's prediction, while plausibility refers to how convincing and understandable the explanation is for humans. It is possible for the explanation to satisfy one of these properties but not the others. For instance, Pruthi et al. \cite{pruthi2020learning} show that attention weights can provide a plausible explanation for human users through a tailored training strategy that hides the fact that the model relies on gender bias to make decisions. This can result in users trusting the model even if it does not work in the way they expect, demonstrating that an unfaithful but plausible explanation can be harmful. Therefore, faithfulness is considered one of the most fundamental properties for an explanation. In this paper, we follow the suggestions from previous works \cite{jacovi2020towards, lyu2022towards} to evaluate explanation methods from the perspective of faithfulness.

Various methods have been proposed to measure the faithfulness of explanations. However, most of these methods are designed for single-step decision-making tasks such as sentiment analysis \cite{deyoung2019eraser} and image classification \cite{yeh2019fidelity}. Few works have been proposed for sequential decision-making tasks. For example, Li et al. \cite{li2020evaluating} propose an evaluation paradigm for neural machine translation that measures the similarity of outputs between the original model and a proxy model trained using only the top-k important tokens as input. However, this paradigm is expensive since a proxy model needs to be retrained for each explanation method. Furthermore, it is based on the assumption that two models will make the same predictions if and only if they use the same reasoning process \cite{jacovi2020towards}. This assumption may not hold, as there exist functionally equivalent models that rely on entirely different reasoning mechanisms, such as different sorting algorithms \cite{lyu2022towards}. Ding et al. \cite{ding2021evaluating} introduce the model consistency test and the input consistency test to measure the faithfulness of explanations. The model consistency test measures the consistency of the output between the distilled model and the original model, based on the same assumption mentioned above. The input consistency test performs substitutions in the input and measures the consistency of feature importance between the original input and the modified input. This test can be used to test the necessary condition of faithfulness but not the sufficient condition \cite{lyu2022towards}. In this paper, we adopt the erasure-based evaluation paradigm proposed by DeYoung et al. \cite{deyoung2019eraser} to measure faithfulness. This paradigm erases parts of the input according to the explanation and observes the change in the output. DeYoung et al. \cite{deyoung2019eraser} apply this paradigm to various single-step decision-making tasks in NLP. And we extend it to VLN, which is a sequential decision-making task.

\section{VLN and Explanation Methods}
\subsection{VLN Models}
Vision-and-Language Navigation (VLN) requires an embodied agent to navigate to the target location in a discrete environment based on natural language instructions.  During the task, the agent receives a natural language instruction $\boldsymbol{U} = \{ u_1, u_2, ..., u_L \}$ consisting of $L$ words. At each time step $t$, the agent perceives a panoramic view $\boldsymbol{V}_t$ and needs to predict an action $a_t$. Specifically, the panoramic view $\boldsymbol{V}_t$ is typically divided into 36 single views $\{\boldsymbol{v}_{t,i}\}^{36}_{i=1}$. In the discrete environment, the agent is provided with a set of $N_t$ navigable locations, whose views from the current location are denoted as $\boldsymbol{C}_t = \{\boldsymbol{c}_{t,i}\}^{N_t}_{i=1}$. Thus, the agent only needs to select the next location from the candidate set as the action $a_t$.

To complete this task, most existing VLN models consist of three processes: language processing, vision processing, and action decision. 

\paragraph{Language Processing.} Each word in the instruction is firstly encoded to a word embedding. Then these embeddings are fed to a RNN or a transformer, which obtains the linguistic features $\boldsymbol{X} = \{ \boldsymbol{x}_1, \boldsymbol{x}_2, ..., \boldsymbol{x}_L \}$.

\paragraph{Vision Processing.} Each view feature $\boldsymbol{f}_{t,i}$ is obtained by concatenating its visual feature encoded by a pre-trained convolutional neural network or a transformer and its orientation feature $\langle \sin \psi_{t,i} ; \cos \psi_{t,i} ; \sin \theta_{t,i} ; \cos \theta_{t,i} \rangle$, where $\psi_{t,i}$ and $\theta_{t,i}$ are the heading and elevation angles, respectively.

\paragraph{Action Decision.} At each timestep $t$, an attention-based model takes linguistic features, view features, and previous state representation as input, and outputs the updated state representation $\boldsymbol{s}_t$ and the action probability $\boldsymbol{p}_t$ among all the candidate actions. Then the agent takes the action $a_t$ which has the maximum probability.

\paragraph{} In this paper, we consider two popular VLN models: Rec-VLN-BERT \cite{hong2021vln}, representing transformer-based models, and EnvDrop \cite{tan2019learning}, representing RNN-based models.

\subsection{Explanation Methods}
The explanation methods that we evaluated in the benchmark can be divided into two categories: attention-based methods and gradient-based methods. In this subsection, we briefly describe each of these methods.

\paragraph{Attention.} Attention mechanisms are widely used in VLN models, and a straightforward way to interpret these models is to utilize the attention weights calculated between the state representation and linguistic features. This approach is known as \textbf{Vanilla Attention} (VaAtt) \cite{clark2019does}. 
In addition, we evaluated an improved version of attention-based explanation method called \textbf{Vector Norms} (VecNorm) \cite{kobayashi2020attention}. VecNorm interprets the model by using the norm of the weighted transformed vector. The importance score of each token is represented by $||{\alpha}_i v(\boldsymbol{x}_i)||_2$, where ${\alpha}$ is the language attention weight of the $i$-th token, and $v(\cdot)$ is the mapping function for the original weighted feature.

\paragraph{Gradient.} Different from attention-based explanation methods restricted to the models with attention mechanisms, gradient-based explanation methods are more general, relying only on the standard backpropagation rules. In this paper, we consider four methods: (1) \textbf{Vanilla Gradient} (VaGrad) \cite{simonyan2013deep}: assigns the importance score of each input token by the $L_2$ norm of its gradient, which can be represented as ${||\frac{\partial \boldsymbol{p}_t[a_t]}{\partial e(u_i)}||}_2$, where $e(\cdot)$ is the word embedding function. (2) \textbf{Gradient $\times$ Input} (GradInp) \cite{denil2014extraction}: computes the importance score by calculating the element-wise product between the input feature and the corresponding gradient, which can be represented as ${\frac{\partial \boldsymbol{p}_t[a_t]}{\partial e(u_i)}} \cdot e(u_i)$. (3) \textbf{Gradient-weighted Class Activation Mapping} (GradCAM) \cite{selvaraju2017grad}: first computes the feature weight by averaging the gradients of features, then calculates the element-wise product between the feature and the corresponding weight to refer the importance score, which can be represented as $(\frac{1}{L} \sum_i \frac{\partial \boldsymbol{p}_t[a_t]}{\boldsymbol{x}_i}) \cdot \boldsymbol{x}_i$. (4) \textbf{Integrated Gradients} (IngGrad) \cite{sundararajan2017axiomatic}: extracts the importance score by computing the element-wise product between the line integral of the gradients from a baseline input $\boldsymbol{U'}$ that does not contain any information for the model prediction to the original input $\boldsymbol{U}$ and the difference of these two inputs in the embedding space.

\section{Evaluation Methodology}

In this section, we describe the pipeline and metrics used to evaluate the faithfulness of explanation methods.

\subsection{Evaluation Pipeline}
An explanation is considered faithful if it can accurately represent the reasoning process behind the model’s prediction. This means that the extracted feature importance from the explanation method should be consistent with the internal decision mechanism of the model. To evaluate the faithfulness of explanation methods, we follow \cite{deyoung2019eraser} to use the erasure-based evaluation pipeline. Specifically, we remove parts of the input features based on the explanation provided by the method. If the most important features (as indicated by the explanation) are first removed, the model's prediction is expected to change drastically. On the other hand, if the most important features are kept and other features are removed, the change in prediction should be smaller.

However, unlike the simple classification task used in \cite{deyoung2019eraser}, VLN is a sequential decision-making task that relies on current observations and historical information. This means that erasing important tokens at a specific time step may not result in missing information for those tokens, as historical information may still contain relevant information. Therefore, we propose a new pipeline to measure faithfulness for the VLN task.

\begin{algorithm}
\caption{Measuring faithfulness at time step $t$.}
\label{alg-comp}
\begin{algorithmic}[1]
    \REQUIRE ~~\\
    Instruction: $\boldsymbol{I}$. \\
    Starting location: ${loc}_1$. \\
    Initial state: $\boldsymbol{s}_0$. \\
    Actions that the agent takes with the instruction $\boldsymbol{I}$ as the textual input: $[a_1, a_2, ..., a_t]$. \\
    Probability among the candidate actions at time step $t$: $\boldsymbol{p}_t$. \\
    Explanation for the decision-making at time step $t$: $\boldsymbol{r}_t$. \\
    \ENSURE ~~\\
    Faithfulness metrics: $\boldsymbol{F}$. \\
    \STATE Get perturbed instruction $\boldsymbol{I'}$ by erasing or preserving top-k important tokens from instruction $\boldsymbol{I}$ according to $\boldsymbol{r}_t$. 
    % \FOR{$i \in [1, t]$ }
    \FOR{$i \gets \text{1}$ \textbf{to} $\text{t}$}
        \STATE $\boldsymbol{V}_i = \text{Env}({loc}_i)$
        \STATE $\boldsymbol{s}_i, \boldsymbol{p}'_i = \text{VLN-Model}(\boldsymbol{V}_i, \boldsymbol{I'}, \boldsymbol{s}_{i-1})$ 
        \IF{ $i < t$}
            \STATE Force the agent to take action $a_i$.
            \STATE The agent reach location ${loc}_{i+1}$.
        \ENDIF
    \ENDFOR
    \STATE $a'_t = \text{argmax}(\boldsymbol{p}'_t)$
    \STATE Calculate $\boldsymbol{F}$ by using $a_t$, $a'_t$, $\boldsymbol{p}_t$, $\boldsymbol{p}'_t$.
    
\end{algorithmic}
\end{algorithm}

As shown in Algorithm \ref{alg-comp}, to measure faithfulness at a specific time step $t$, we need to erase or preserve important tokens at the beginning. At each step, the VLN model takes the observation, previous state, and the perturbed instruction as input and outputs the updated state and action probability. However, the selection of actions is not determined solely by the maximum probability assigned to them by the agent. Instead, we force the agent to take the original action that it takes when its textual input is the original instruction. This ensures that the agent has the same visual observations as the original agent. Therefore, the difference between the output and the original output at time step $t$ is caused by the erased tokens. Finally, we can utilize actions and action probabilities to calculate the faithfulness metrics.

\subsection{Faithfulness Metrics}
We adopt four metrics to measure faithfulness. 

\paragraph{Comprehensive} (COMP) \cite{deyoung2019eraser} measures the faithfulness by the change of the output probability of the original predicted action after the top-k important tokens are erased, which can be defined as follows:
\begin{equation}
    \text{COMP} = \boldsymbol{p}_t(\boldsymbol{I})[a_t] - \boldsymbol{p}_t(\boldsymbol{I} \backslash \boldsymbol{I}_{:k})[a_t].
\end{equation}

\paragraph{Sufficiency} (SUFF) \cite{deyoung2019eraser} only preserves the top-k important tokens and computes the change in output probability of the original predicted action to represent the faithfulness:
\begin{equation}
    \text{SUFF} = \boldsymbol{p}_t(\boldsymbol{I})[a_t] - \boldsymbol{p}_t(\boldsymbol{I}_{:k})[a_t].
\end{equation}

\paragraph{Decision Flip without Rationale} (DFw/oR) \footnote{DFw/oR is a simple extension of Decision Flip - Most Informative Token \cite{chrysostomou2021improving} which only measures the rate of decision flip subsequent to the removal of the most important (top-1) token.} assumes that the explanation is faithful if the predicted action is changed after erasing the top-k important tokens:
\begin{equation}
    \text{DFw/oR}= \begin{cases}1 & \text { if } a_t(\boldsymbol{I}) \neq a_t\left(\boldsymbol{I} \backslash \boldsymbol{I}_{: k}\right) \\ 0 & \text { if } a_t(\boldsymbol{I})=a_t\left(\boldsymbol{I} \backslash \boldsymbol{I}_{: k}\right)\end{cases}
\end{equation}

\paragraph{Decision Flip with Rationale} (DFw/R) assumes that the explanation is unfaithful if the predicted action is changed after only preserving the top-k important tokens:
\begin{equation}
    \text{DFw/R}= \begin{cases}1 & \text { if } a_t(\boldsymbol{I}) \neq a_t\left(\boldsymbol{I}_{: k}\right) \\ 0 & \text { if } a_t(\boldsymbol{I})=a_t\left(\boldsymbol{I}_{: k}\right)\end{cases}
\end{equation}

COMP and DFw/oR can be used to evaluate whether the tokens assigned lower weights are unnecessary for the predictions. On the other hand, SUFF and DFw/R can be used to measure whether the tokens assigned high weights are sufficient for a model to make the original decision.

\section{Experiments}

\subsection{Setup}
\paragraph{Datasets.} We adopt two popular VLN datasets, R2R \cite{Anderson_2018_CVPR} and RxR \cite{rxr}, to evaluate the faithfulness of explanation methods. Both datasets are built upon the Matterport3D dataset \cite{chang2017matterport3d} and contain step-by-step navigation instructions. The RxR dataset has more detailed instructions than R2R. According to whether the environment has appeared in the training set, the validation set of these two datasets can be divided into two parts: validation-seen (val-seen) and validation-unseen (val-unseen). The val-seen set of R2R contains 1020 path-instruction pairs, and the val-unseen set contains 2349 path-instruction pairs. In these two validation sets, the average length of instructions is 30 tokens, and the average length of expert paths is 6 steps. For the RxR dataset, we use its English subset, which includes 2939 path-instruction pairs in the val-seen set and 4551 path-instruction pairs in the val-unseen set. In both validation sets, the average length of instructions is 113 tokens, and the average length of expert paths is 9 steps.

\begin{table}
\centering
\begin{tabular}{l|cc|cc}
\toprule
 \multicolumn{1}{l|}{} & \multicolumn{2}{c|}{R2R (Val-Seen)}                      & \multicolumn{2}{c}{R2R (Val-Unseen)}                      \\
 & SR     & SPL     & SR    & SPL  \\
\hline
Rec-VLN-BERT & 72.2\%   & 67.7\%    & 62.8\%    & 56.8\%  \\
EnvDrop      & 66.5\%  & 63.6\%   & 52.1\%   & 48.6\% \\
\bottomrule
\end{tabular}
\caption{Navigation performance on the R2R dataset. SR refers to Success Rate, which is the ratio of stopping within 3 meters to the target location. SPL refers to Success rate weighted by Path Length, which measures both the accuracy and efficiency of navigation. }
\label{r2r_nav}
\end{table}

\begin{table}
\centering
\begin{tabular}{l|cc|cc}
\toprule
 \multicolumn{1}{l|}{} & \multicolumn{2}{c|}{RxR (Val-Seen)}                      & \multicolumn{2}{c}{RxR (Val-Unseen)}                      \\
 & SR     & SPL     & SR    & SPL  \\
\hline
Rec-VLN-BERT & 49.9\%   & 45.5\%    & 45.6\%    & 40.1\%  \\
EnvDrop      & 42.6\%  & 39.5\%   & 35.7\%   & 31.5\% \\
\bottomrule
\end{tabular}
\caption{Navigation performance on the RxR dataset. }
\label{rxr_nav}
\end{table}

\begin{table*}[t!]
\centering
\begin{tabular}{l|cccc|cccc}
\toprule
\multicolumn{1}{l|}{} & \multicolumn{4}{c|}{R2R (Val-Seen)}                                               & \multicolumn{4}{c}{R2R (Val-Unseen)}                                               \\
\multicolumn{1}{l|}{} & DFw/oR$\uparrow$ &  COMP$\uparrow$ & DFw/R$\downarrow$ & SUFF$\downarrow$ & DFw/oR$\uparrow$ &  COMP$\uparrow$ & DFw/R$\downarrow$ & SUFF$\downarrow$ \\
\hline
\textbf{Rec-VLN-BERT}         &               &                   &               &                   &               &                   &               &                   \\
% Random                        &               &                   &               &                   &               &                   &               &                   \\
VaAtt                & 26.7\% & 17.4\%        & 43.1\% & 32.1\%            & 28.7\% & 18.5\%        & 47.0\% & 35.0\%            \\
VecNorm              & 27.2\% & 17.9\%        & 42.7\% & 31.4\%            & 29.4\% & 19.2\%        & 46.0\% & 34.1\%            \\
VaGrad             & 33.7\% & 22.5\%        & 37.3\% & 27.0\%            & 36.3\% & 24.5\%        & 41.0\% & 29.7\%            \\
GradInp           & 33.0\% & 22.1\%        & 38.3\% & 28.0\%            & 35.4\% & 23.8\%        & 42.0\% & 30.5\%            \\
GradCAM             & 33.8\% & 22.6\%        & 37.8\% & 27.6\%            & 36.0\% & 24.3\%        & 40.8\% & 29.8\%            \\
IngGrad                   & \textbf{44.2\%} & \textbf{31.5\%}        & \textbf{28.5\%} & \textbf{19.6\%}            & \textbf{48.1\%} & \textbf{34.3\%}        & \textbf{31.1\%} & \textbf{21.3\%}            \\
% Random                        &               &                   &               &                   &               &                   &               &                   \\
% Human  &               &                   &               &                   &               &                   & \\
\hline
\textbf{EnvDrop}              &               &                   &               &                   &               &                   &               &                   \\
% Random                        &               &                   &               &                   &               &                   &               &                   \\
VaAtt                & 18.4\% & 9.4\%         & 43.8\% & 29.6\%            & 21.3\% & 10.1\%        & 49.8\% & 32.7\%            \\
VecNorm              & 24.7\% & 14.1\%        & 39.5\% & 25.6\%            & 26.7\% & 14.4\%        & 46.0\% & 29.2\%            \\
VaGrad             & 21.8\% & 12.6\%        & 41.3\% & 27.4\%            & 25.1\% & 13.7\%        & 47.7\% & 30.7\%            \\
GradInp           & \textbf{42.1\%} & \textbf{28.8\%}        & 31.7\% & 18.6\%            & 47.5\% & 30.9\%        & 37.5\% & 21.6\%            \\
GradCAM             & 33.5\% & 21.8\%        & 32.9\% & 19.6\%            & 38.6\% & 23.8\%        & 38.9\% & 22.6\%            \\
IngGrad                   & 41.8\% & 27.8\%        & \textbf{30.5\%} & \textbf{17.8\%}            & \textbf{49.4\%} & \textbf{32.2\%}        & \textbf{35.8\%} & \textbf{20.2\%} \\
% Human  &               &                   &               &                   &               &                   & \\
\bottomrule
\end{tabular}
\caption{Faithfulness benchmark result on the R2R dataset.}
\label{r2r_result}
\end{table*}

\begin{table*}[t!]
\centering
\begin{tabular}{l|cccc|cccc}
\toprule
\multicolumn{1}{l|}{} & \multicolumn{4}{c|}{RxR (Val-Seen)}                                               & \multicolumn{4}{c}{RxR (Val-Unseen)}                                               \\
\multicolumn{1}{l|}{} & DFw/oR$\uparrow$ &  COMP$\uparrow$ & DFw/R$\downarrow$ & SUFF$\downarrow$ & DFw/oR$\uparrow$ &  COMP$\uparrow$ & DFw/R$\downarrow$ & SUFF$\downarrow$ \\
\hline
\textbf{Rec-VLN-BERT}         &               &                   &               &                   &               &                   &               &                   \\
% Random                        &               &                   &               &                   &               &                   &               &                   \\
VaAtt                & 43.1\% & 31.6\%        & 40.3\% & 28.7\%            & 47.2\% & 33.8\%        & 43.8\% & 30.5\%            \\
VecNorm              & 45.0\% & 33.3\%        & 38.5\% & 27.3\%            & 49.7\% & 35.9\%        & 41.4\% & 28.4\%            \\
VaGrad             & 50.1\% & 37.7\%        & 30.9\% & 20.8\%            & 54.5\% & 40.0\%        & 34.6\% & 22.9\%            \\
GradInp           & 41.8\% & 30.3\%        & 38.5\% & 27.4\%            & 46.9\% & 33.3\%        & 41.4\% & 28.5\%            \\
GradCAM             & 45.9\% & 33.4\%        & 40.1\% & 28.9\%            & 50.1\% & 35.4\%        & 42.7\% & 29.9\%            \\
IngGrad             & \textbf{70.0\%} & \textbf{55.0\%}        & \textbf{20.5\%} & \textbf{10.5\%}            & \textbf{73.4\%} & \textbf{56.4\%}        & \textbf{23.4\%} & \textbf{12.0\%}            \\
% Random                        &               &                   &               &                   &               &                   &               &                   \\
% Human  &               &                   &               &                   &               &                   & \\
\hline
\textbf{EnvDrop}              &               &                   &               &                   &               &                   &               &                   \\
% Random                        &               &                   &               &                   &               &                   &               &                   \\
VaAtt                & 32.9\% & 19.9\%        & 43.8\% & 28.2\%            & 38.5\% & 22.0\%        & 51.0\% & 31.7\%            \\
VecNorm              & 36.8\% & 23.3\%        & 38.6\% & 23.6\%            & 43.2\% & 26.0\%        & 44.8\% & 26.3\%            \\
VaGrad             & 41.2\% & 27.4\%        & 34.4\% & 19.6\%            & 48.3\% & 30.6\%        & 39.8\% & 21.6\%            \\
GradInp           & \textbf{67.8\%} & \textbf{48.4\%}        & 22.4\% & 8.2\%             & \textbf{76.2\%} & \textbf{52.4\%}        & 27.0\% & 9.2\%             \\
GradCAM             & 59.6\% & 41.4\%        & 24.0\% & 9.2\%             & 67.9\% & 45.5\%        & 28.3\% & 10.1\%            \\
IngGrad              & 63.6\% & 45.5\%        & \textbf{20.1\%} & \textbf{6.4\%}             & 72.0\% & 49.9\%        & \textbf{23.9\%} & \textbf{7.0\%} \\
% Human  &               &                   &               &                   &               &                   & \\
\bottomrule
\end{tabular}
\caption{Faithfulness benchmark result on the RxR dataset.}
\label{rxr_result}
\end{table*}

\paragraph{VLN Models.} For the R2R dataset, we implement Rec-VLN-BERT and EnvDrop by following the open-source codes \footnote{Rec-VLN-BERT: https://github.com/YicongHong/Recurrent-VLN-BERT; EnvDrop: https://github.com/airsplay/R2R-EnvDrop.}. For the RxR dataset, we make adjustments to the path length and instruction length. Specifically, we set the instruction length as 160 for EnvDrop. For Rec-VLN-BERT, we set the path length as 20 and the instruction length as 160.

For the R2R dataset, we use the available model weights of Rec-VLN-BERT and train EnvDrop by ourselves. The navigation performance of the two models is shown in Table \ref{r2r_nav}. For the RxR dataset, due to the lack of available model weights for these two models, we train them by ourselves. Their navigation performance is shown in Table \ref{rxr_nav}.

\paragraph{Explanation Methods.} For Vanilla Attention (VaAtt), since there are several cross-attention modules in Rec-VLN-BERT, we follow \cite{lin2022adapt, cheng2022learning} to use the attention weights obtained in the last module. For Vector Norms (VecNorm), it is designed for transformers. We extend it to general attention modules which have a fully connected layer to transform the attention-weighted features. For Integrated Gradients (IngGrad), we use step size $N=20$ to explain the Rec-VLN-BERT Model in the RxR dataset, due to the limited memory of GPU. The default value of step size is 100. And we consider an empty instruction filled with padding tokens as the baseline input. For other explanation methods, there are no special settings.

\paragraph{Evaluation Methodology.} Considering out-of-distribution (OOD) problems in the erasure-based evaluation \cite{hooker2019benchmark}, we adopt two erasure operations as suggested by \cite{hase2021out}. (1) \textbf{Slice Out} \cite{deyoung2019eraser}, which removes selected tokens from the original instruction, resulting in a shorter length of the perturbed instruction. (2) \textbf{Token Replacement}, which replaces the selected tokens with MASK or UNK tokens for transformer-based or RNN-based models, respectively. We report the results averaged over these two erasure operations.

The number of tokens to erase is a hyper-parameter. We follow \cite{deyoung2019eraser} to set it to the average rationale length provided by humans. LA-R2R \cite{cheng2022learning} and Landmark-RxR \cite{he2021landmark} provide human-annotated rationales of step-wise decision-making for the R2R and RxR datasets, respectively. For the R2R dataset, the average human-annotated rationale length is 4 tokens for both the val-seen and val-unseen sets. For the RxR dataset, the average human-annotated rationale length is 25 tokens and 26 tokens for the val-seen and val-unseen sets, respectively. 

\begin{table*}[t!]
\centering
\begin{tabular}{l|cccc|cccc}
\toprule
\multicolumn{1}{l|}{} & \multicolumn{4}{c|}{R2R (Val-Seen)}                                               & \multicolumn{4}{c}{R2R (Val-Unseen)}                                               \\
\multicolumn{1}{l|}{} & DFw/oR$\uparrow$ &  COMP$\uparrow$ & DFw/R$\downarrow$ & SUFF$\downarrow$ & DFw/oR$\uparrow$ &  COMP$\uparrow$ & DFw/R$\downarrow$ & SUFF$\downarrow$ \\
\hline
\textbf{Rec-VLN-BERT}         &               &                   &               &                   &               &                   &               &                   \\
IngGrad              & \textbf{45.4\%} & \textbf{32.2\%}        & \textbf{30.0\%} & \textbf{20.3\%}            & \textbf{49.1\%} & \textbf{34.9\%}       & \textbf{31.8\%} & \textbf{21.8\%}            \\
Human                & 34.8\% & 23.5\%        & 36.3\% & 25.6\%            & 36.7\% & 24.7\%        & 39.8\% & 28.7\%            \\
\hline
\textbf{EnvDrop}              &               &                   &               &                   &               &                   &               &                   \\
IngGrad              & \textbf{42.0\%} & \textbf{27.6\%}        & \textbf{30.4\%} & \textbf{17.8\%}           & \textbf{50.2\%} & \textbf{31.9\%}       & \textbf{34.0\%} & \textbf{18.1\%}            \\
Human                & 25.3\% & 15.0\%        & 35.6\% & 22.4\%            & 28.3\% & 15.5\%        & 41.5\% & 24.4\% \\
\bottomrule
\end{tabular}
\caption{Comparison result between explanations from IngGrad and human-annotated rationales on the R2R dataset in terms of faithfulness.}
\label{r2r_comparison}
\end{table*}

\subsection{Main Results}

Table~\ref{r2r_result} presents the result of our faithfulness benchmark on the R2R dataset. Among the attention-based explanation methods, we can find that VecNorm outperforms VaAtt in explaining both Rec-VLN-BERT and EnvDrop, indicating the effectiveness of our extension of VecNorm for EnvDrop. On the other hand, for gradient-based explanation methods, we can observe that they are generally more faithful than attention-based methods, except VaGrad which performs worse than VecNorm in explaining EnvDrop.

Regarding the performance of different explanation methods on Rec-VLN-BERT, we can find that IngGrad achieves the highest faithfulness score among all metrics, with a significant lead over other methods. For EnvDrop, the performance of IngGrad and GradInp are comparable. 

Table~\ref{rxr_result} displays the result of our faithfulness benchmark on the RxR dataset. Similar to the findings on the R2R dataset, IngGrad is the most effective method in explaining Rec-VLN-BERT. In addition, we also observe that IngGrad and GradInp perform comparably in explaining EnvDrop. Notably, the performance of the same explanation method on the RxR dataset is significantly better than on the R2R dataset. This can be attributed to the difference in erased token rates between the two datasets. The erased token rate on the RxR dataset is 23\%, which is nearly twice as high as the rate on the R2R dataset (13\%).

\subsection{Comparison with Human Explanations}

% Do the VLN models make decisions in the same way as humans?
From these benchmarks mentioned above, we can observe that IngGrad performed well in explaining both Rec-VLN-BERT and EnvDrop. And we currently have human-annotated rationales for decision-making in expert paths. We further compare the explanations derived from IngGrad with human-annotated rationales to investigate which can explain the VLN models better. For a fair comparison, we only measure the faithfulness on the steps whose locations appear in both expert paths and predicted paths. And at each valid step, we set the number of tokens to erase to the length of human-annotated rationale. Based on these settings, we conduct experiments on the R2R dataset.

The comparison result is shown in Table \ref{r2r_comparison}, and it reveals that IngGrad's explanations are more faithful than the human-annotated rationales in interpreting Rec-VLN-BERT and EnvDrop on both val-seen and val-unseen sets. Notably, the DFw/R values of the human-annotated rationales are worth considering. With the human-annotated rationales kept, only 36.3\% and 39.8\% of decisions made by Rec-VLN-BERT are changed on the val-seen and val-unseen sets, respectively. These values indicate that the human-annotated rationales can support almost 62\% of the decisions made by Rec-VLN-BERT, showing a degree of consistency between the inner reasoning process of Rec-VLN-BERT and human explanations. The DFw/R values of the human-annotated rationales for EnvDrop are similar. With the human-annotated rationales kept, 35.6\% and 41.5\% of decisions made by EnvDrop are changed on the val-seen and val-unseen sets, respectively, indicating a degree of consistency between the inner reasoning process of EnvDrop and human explanations.

However, the human-annotated rationales achieve only 34.8\% DFw/oR on the val-seen set and 36.7\% DFw/oR on the val-unseen set for Rec-VLN-BERT. This suggests that Rec-VLN-BERT leverages other tokens apart from the human-annotated rationale to support nearly 65\% of its decisions, indicating a difference between the inner reasoning process of Rec-VLN-BERT and that of humans. For EnvDrop, the DFw/oR values of the human-annotated rationales are even worse. Without the human-annotated rationales, only 25.3\% and 28.3\% of decisions made by EnvDrop are changed on the val-seen and val-unseen sets, respectively, indicating a greater difference between the inner reasoning process of EnvDrop and human explanations compared to Rec-VLN-BERT. Overall, these results suggest that while the internal reasoning process of these models may be somewhat similar to human explanations, they still exhibit a degree of difference from human explanations.

\subsection{Visualization}
\begin{figure*}[t]
\centering
\includegraphics[width=0.98\textwidth]{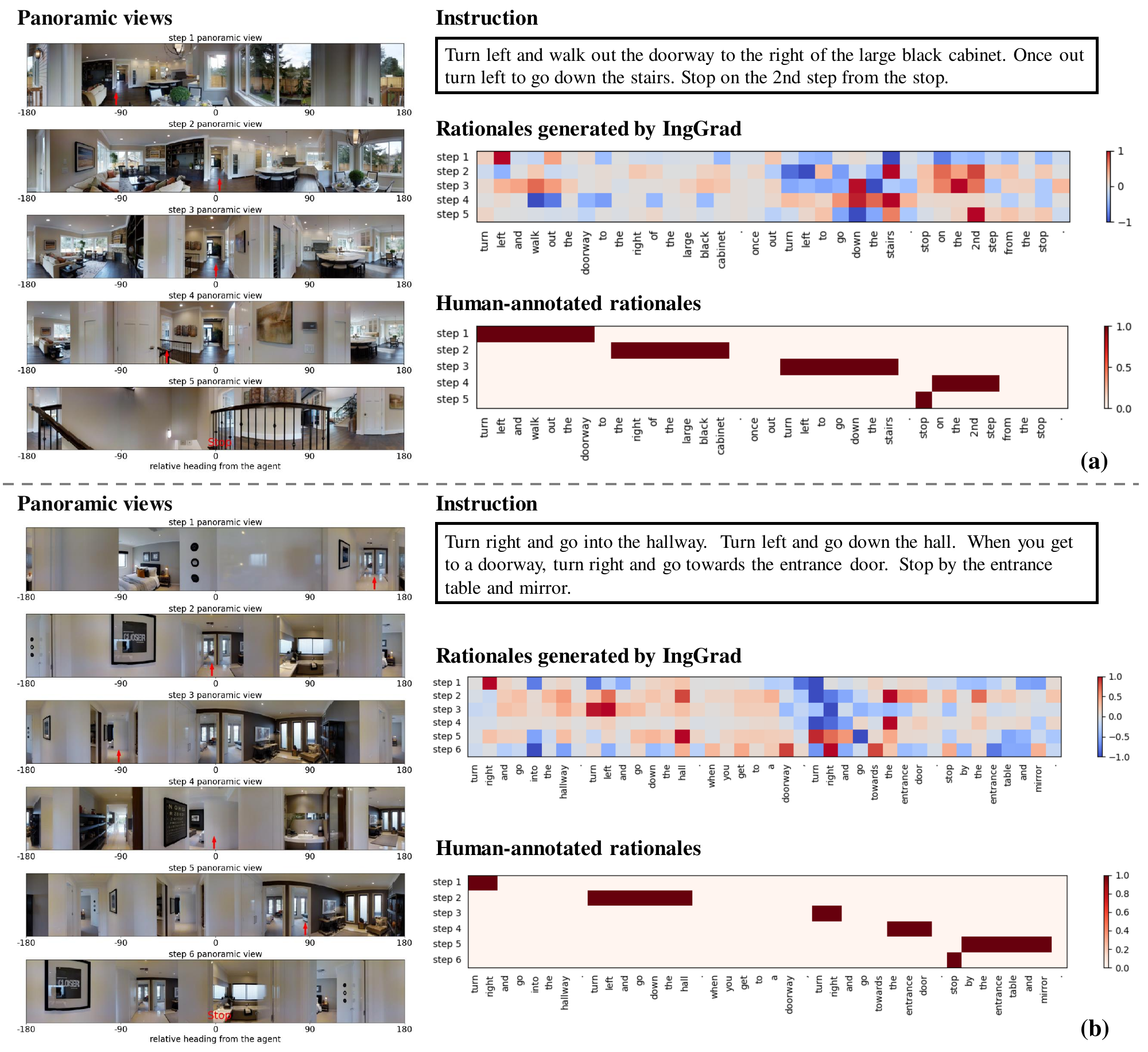}
\caption{Visualization of two samples in R2R val-unseen set. Each red arrow in the panoramic view represents the direction to the next step. In the rationales generated by IngGrad, the patches of red and blue represent positive and negative importance scores, respectively. In the human-annotated rationales, the red patches indicate that the corresponding tokens are important. } \label{fig_case}
\end{figure*}

In Fig.~\ref{fig_case}, we visualize two samples to demonstrate the difference between explanations generated by the best explanation method in our benchmark (i.e., IngGrad) and the human annotator. The IngGrad explanation significantly differs from the human explanation at the second and fourth steps in Fig.~\ref{fig_case}(a). Interestingly, retaining all the important tokens identified by the human annotator prompts the VLN model to alter its decision at these points. Conversely, preserving the same number of important tokens indicated by IngGrad supports the VLN model in maintaining its original decision. Similar differences occur at the third and fifth steps in Fig.~\ref{fig_case}(b). These results show that, in these cases, IngGrad-generated explanations are more faithful than those provided by human annotators.

\subsection{Discussion}
In this subsection, we discuss the key findings revealed by the experimental results mentioned above.

(1) IngGrad's notable success in explaining both two VLN models can be attributed to its approach in addressing the gradient saturation problem. By integrating gradients during the scaling up of inputs from a starting value to their current value, IngGrad effectively captures crucial gradient-related information. This leads to the generation of accurate and meaningful explanations. As for the competitive performance of GradInp in explaining EnvDrop, we still need to conduct more empirical exploration to find out the reason in the future.

(2) On the other hand, attention-based methods exhibit poor performance in explaining both VLN models. As highlighted by \cite{bai2021attentions}, the calculation of attention is influenced by biased input, which often results in the allocation of attention to uninteresting tokens. Consequently, the explanations generated by attention-based methods may lack relevance and coherence.

(3) Moreover, the internal reasoning processes of the two VLN models appear to differ from human explanations when making similar decisions. One possible reason is that these models lack direct and precise guidance on which words to rely upon during the training process. To achieve better alignment between human and agent explanations, it is crucial to provide explicit and accurate guidance for model learning.

\section{Conclusions}
In this paper, we propose a new pipeline to effectively evaluate the faithfulness of the explanation method based on erasure, which provides a quantitative comparison among six different explanation methods on two widely used VLN datasets. To the best of our knowledge, it is the first work to evaluate explanation methods for vision-and-language navigation. Through the benchmarks we presented, we find that IngGrad is the best method to explain Rec-VLN-BERT (a representative transformer-based model) among the six explanation methods we examined. And both IngGrad and GradInp are good choices to explain EnvDrop (a representative RNN-based model). However, it is worth noting that the attention map does not faithfully explain the decision-making processes of either of these representative VLN models.
Besides, we also observe that the internal reasoning processes of two representative VLN models exhibit a degree of difference from human explanations through the analysis experiment. There is still a large room for improvement in the alignment between agents and humans.

With regard to the limitations of our work, our benchmark examined two attention-based and four gradient-based methods, but not other types of explanation methods, such as propagation-based methods \cite{bach2015pixel, voita2019analyzing}. However, our proposed evaluation paradigm is applicable to assess any method that generates importance scores as explanations. In the future, we aim to use this paradigm to evaluate additional algorithms for explaining VLN models. Furthermore, as VLN models rely on both textual and visual cues to make decisions, evaluating the faithfulness of the visual explanations generated by the explanation methods remains an underexplored territory in our research. Therefore, delving into this aspect in future studies will provide a promising avenue to gain a comprehensive understanding of VLN models in a multi-modal context.

% \section{Acknowledgements}
\ack This research is partly supported by the Innovation and Technology Commission of the HKSAR Government through the InnoHK initiative. Jia Pan is also partially supported by ITF GHP/126/21GD.

% \bibliography{ecai}

\end{document}